\documentclass[11pt, a4paper]{thuc3i}
\usepackage[sort&compress]{natbib}
\setheadertext{Self-Manager: Parallel Agent Loop for Long-form Deep Research}

\usepackage[utf8]{inputenc} 
\usepackage[T1]{fontenc}    
\usepackage{hyperref}       
\usepackage{url}            
\usepackage{tabularx}
\usepackage{makecell}
\usepackage{array,ragged2e,booktabs}
\newcolumntype{P}[1]{>{\RaggedRight\arraybackslash}p{#1}}
\usepackage{longtable}
\usepackage{amsfonts}       
\usepackage{amsthm}         

\usepackage{nicefrac}       
\usepackage{microtype}      
\usepackage{xcolor}         

\usepackage{algorithm}
\usepackage{algorithmicx}
\usepackage{algpseudocode}

\definecolor{darkblue}{rgb}{0, 0, 0.5}
\definecolor{darkred}{rgb}{0.5, 0, 0}
\hypersetup{colorlinks=true, citecolor=darkred, linkcolor=darkred, urlcolor=darkred}

\usepackage{xurl}

\usepackage{soul}
\usepackage{amsmath}
\usepackage{subcaption}
\usepackage{graphicx}
\usepackage{changes}
\usepackage{mathtools}
\usepackage{changes}
\usepackage{pifont}
\usepackage{tocloft}

\usepackage{lipsum}
\usepackage{colortbl}
\usepackage{wrapfig}
\usepackage{xspace}
\usepackage{multirow}
\usepackage{enumitem}

\usepackage{pifont}
\usepackage{listings}
\usepackage{fontawesome5} 

\usepackage{caption}

\usepackage{makecell}

\usepackage{tabularx}
\usepackage{afterpage}

\usepackage[edges]{forest}
\usepackage[normalem]{ulem}
\usepackage{caption}
\usepackage{CJKutf8}
\usepackage{bbding}
\usepackage[most,skins,theorems]{tcolorbox}
\usepackage{epigraph}

\usepackage[tikz]{bclogo}
\usepackage[framemethod=tikz]{mdframed}

\definecolor{darkblue}{rgb}{0, 0, 0.5}
\hypersetup{colorlinks=true, citecolor=darkred, linkcolor=darkred, urlcolor=darkred}

\usepackage{arydshln}
\definecolor{lightcyan}{RGB}{224, 255, 255}
\definecolor{lightpink}{RGB}{215,246,255}
\definecolor{lightyellow}{rgb}{1.0, 1.0, 0.8}
\definecolor{lightblue}{rgb}{0.8, 0.9, 1.0}
\definecolor{algorithm}{RGB}{251,234,255}
\definecolor{mygreen}{RGB}{0,120,90}
\definecolor{lightorange}{HTML}{faa755}
\usepackage[most,skins,theorems]{tcolorbox}

\usepackage{xcolor}

\usepackage{xspace}

\usepackage{cleveref}

\usepackage{enumitem}

\usepackage[dvipsnames]{xcolor}

\def\github{\raisebox{-1.5pt}{\includegraphics[height=1.05em]{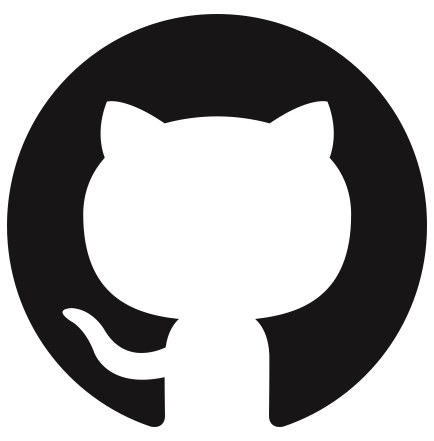}}}

\title{Self-Manager: Parallel Agent Loop for Long-form Deep Research}

\author{%
    Yilong Xu$^{1, 2}$, Zhi Zheng$^{2}$, Xiang Long$^{2}$, Yujun Cai$^{3}$, Yiwei Wang$^{4}$ \\
    $^{1}$University of Chinese Academy of Sciences \quad
    $^{2}$ModelBest Inc. \quad
    $^{3}$University of Queensland \quad
    $^{4}$University of California, Merced\\
    \vskip1mm
    \faEnvelope[regular]~\texttt{yiloxuu@gmail.com}  \\
    \vskip2mm
    \github \quad \url{https://github.com/ylXuu/ParallelAgent}
}

\begin{abstract}
Long-form deep research requires multi-faceted investigations over extended horizons to get a comprehensive report. When handling such complex tasks, existing agents manage context at the subtask level to overcome linear context accumulation and information loss. However, they still adhere to a single context window and sequential execution paradigm, which results in mutual interference and blocking behavior, restricting scalability and adaptability. To address this issue, this paper introduces Self-Manager, a parallel agent loop that enables asynchronous and concurrent execution. The main thread can create multiple subthreads, each with its own isolated context, and manage them iteratively through Thread Control Blocks, allowing for more focused and flexible parallel agent execution. To assess its effectiveness, we benchmark Self-Manager on DeepResearch Bench, where it consistently outperforms existing single-agent loop baselines across all metrics. Furthermore, we conduct extensive analytical experiments to demonstrate the necessity of Self-Manager’s design choices, as well as its advantages in contextual capacity, efficiency, and generalization.
\end{abstract}

\begin{document}

\maketitle

\section{Introduction}

Language agents \citep{su-etal-2024-language} are directing AI skills toward tasks that are substantially more challenging for humans, such as Deep Research \citep{du2025deepresearchbenchcomprehensivebenchmark, wei2025browsecompsimplechallengingbenchmark}. Current solutions for such complicated workloads generally rely on multi-agent systems constructed with predefined workflows. However, these systems suffer from limited generalization across scenarios and high deployment overhead. In contrast, the single-agent paradigm \citep{jin2025searchr1trainingllmsreason, nguyen2025sfrdeepresearcheffectivereinforcementlearning}, which is free from predetermined workflows and capable of adaptive iteration via a basic and generic loop, provides a more flexible, robust, and ultimately promising agent architecture.

Existing single-agent loops mostly employ ReAct's "Think–Act–Observe" pattern \citep{yao2022react}. However, this classical design is bound by linearly accumulated context, causing salient information to be readily diluted and incapable of sustaining long-horizon execution \citep{gao2025turnsunlockinglonghorizonagentic}. To solve this issue, several approaches allow the agent to summarize the prefix context during the act phase \citep{wu2025resumunlockinglonghorizonsearch, nguyen2025sfrdeepresearcheffectivereinforcementlearning}. Although this design reduces the rigidity of linear context growth, improper condensation timing may result in severe information loss. At a fundamental level, the above paradigms are constrained, as they neglect the critical feature of context management granularity when dealing with complex tasks.

Heuristically, a \textit{subtask} can be treated as a natural unit for agentic context management, as decomposing complex tasks into subtasks often yields more effective solutions. \citet{schroeder-etal-2025-thread, ye2025agentfoldlonghorizonwebagents, sun2025scaling} adopt the idea of branching subtasks, compressing each subtask's working memory through mechanisms such as folding. However, these methods share a common limitation: subtasks are executed sequentially within a shared context window. This causes two issues: (1) contexts of different task levels interfere with each other, and (2) the global task remains blocked until the currently running subtask is completed. Both limit the scalability and flexibility of tackling complicated problems at the subtask granularity.

\begin{figure*}[t]
    \centering
    \includegraphics[width=\linewidth]{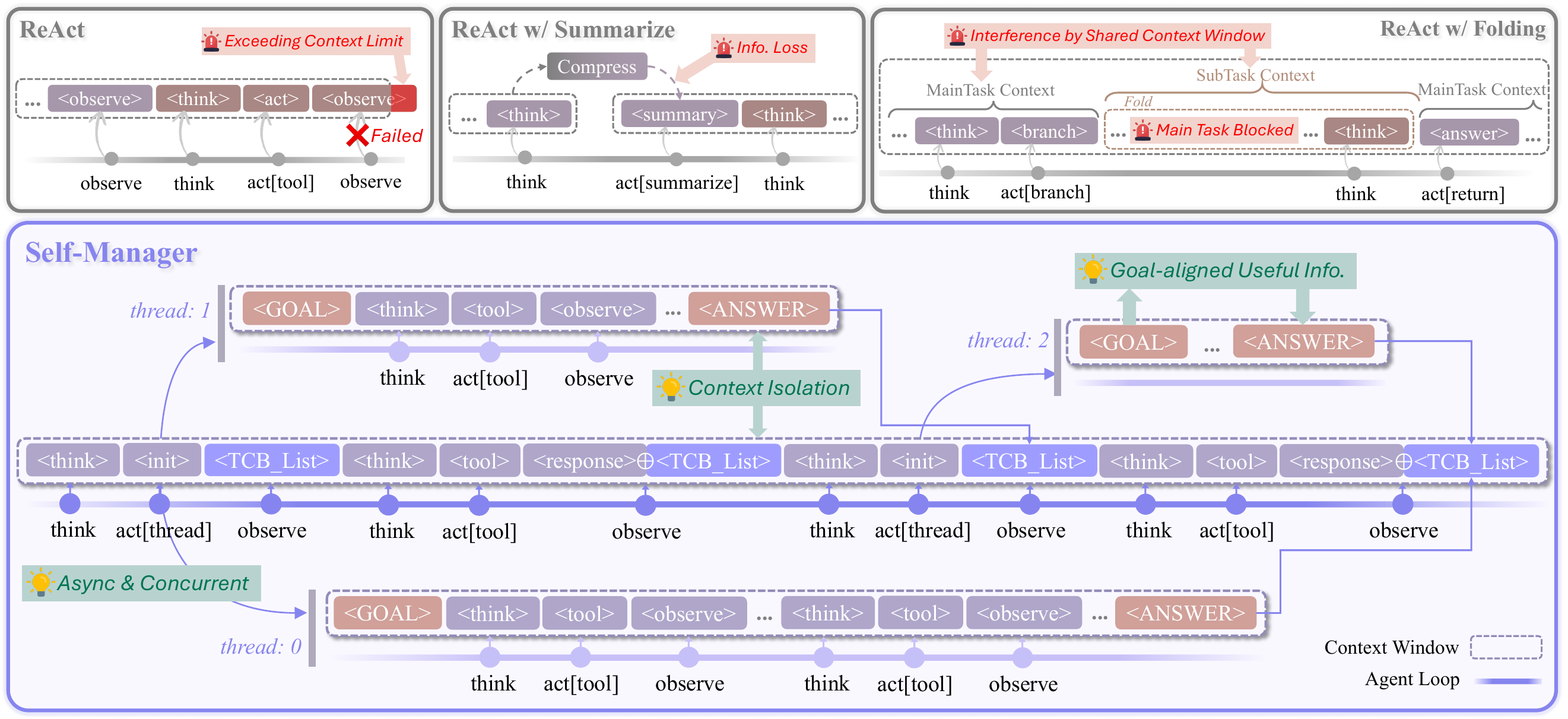}
    \caption{Overview of our proposed Self-Manager. Compared to classical agent loops, Self-Manager spawns subthreads asynchronously and concurrently, with contexts isolated between threads. Upon completion, each subthread returns its result to the main thread's most recent observation. During execution, the main thread can operate on subthreads and monitor their latest states via the TCB list, thereby enabling autonomous management.}
    \label{fig:teaser}
\end{figure*}

In this paper, we propose \textbf{Self-Manager}, which is a parallel single-agent loop architecture. Specifically, we introduce the concept of \textit{threads}, where the main thread directly handles the user task and can spawn subthreads. This allows for the decomposition of complex tasks and the distribution of subtasks to subthreads. Notably, Self-Manager exhibits the following features:
\begin{itemize}
    \item \textbf{Asynchrony}. Self-Manager executes in parallel, and the main thread is not blocked by the execution of subthreads.
    \item \textbf{Concurrency}. Self-Manager can initialize multiple subthreads simultaneously at any iteration turn, enabling parallel exploration of different perspectives of a task.
    \item \textbf{Context Isolation}. In Self-Manager, each thread has its own context window, ensuring that subtasks do not interfere with one another.
\end{itemize}
To enable these capabilities in the parallel agent loop, we introduce the \textit{Thread Control Block} (TCB), a structured object encapsulating each subthread's information. The main threads can monitor all subthreads by observing the TCB list, thereby enabling self-managed multi-threaded execution.

We conduct experiments on DeepResearch Bench \citep{du2025deepresearchbenchcomprehensivebenchmark}, a challenging benchmark for long-form deep research. The results show that Self-Manager achieves the best performance compared with other single-agent loop paradigms, and narrows the performance gap with proprietary deep research systems such as Gemini Deep Research \citep{gemini_deep_research}. Furthermore, we examine the component necessity, contextual capabilities, efficiency, and generalizability of Self-Manager, providing a multi-dimensional analysis of the improvements offered by Self-Manager as a parallel agent loop.

Our main contributions are as follows: (1) We propose a novel parallel agent loop, Self-Manager, capable of asynchronously and concurrently conducting tool-integrated inference to handle complex tasks. (2) We design the Thread Control Block and incorporate it into the agent loop to enable self-management of the parallel execution in Self-Manager, hence ensuring flexibility. (3) Our proposed Self-Manager outperforms single-agent baselines on DeepResearch Bench benchmark.

\section{Methodology}

In this section, we provide a detailed description of the modeling of Self-Manager.

\subsection{Background}

We consider an agent following the ReAct paradigm \citep{yao2022react}, which interleaves reasoning and environment interaction through a sequence of actions and observations. A standard ReAct agent loop trajectory of length $T$ is:
\begin{equation}
    \tau = \left \{ a_1, o_1, a_2, o_2, \dots , a_T, o_T, \mathcal{A} \right \},
\end{equation}
where the observation of environment $o_t$ is from the action $a_t$ sampled from the agent's policy $a_{t} \sim \pi_{\theta}\left(\cdot \mid h_{t-1}\right)$, with $h_{t-1}$ denoting the historical context accumulated so far. In practice, each step’s action can decompose into a thinking and an acting component: $a_t \longrightarrow \left \{ a_t^{\mathrm{think}}, a_t^{\mathrm{act}} \right \} $. This generic loop is, in principle, applicable to a wide range of tasks. However, the linear context growth prevents it from supporting long-horizon execution.

Although some prior work compresses historical context when approaching the context limit, such compression can easily lead to the loss of critical information. Thus, methods such as FoldAgent \citep{sun2025scaling} manage context at the granularity of subtasks. While this design better accommodates complex tasks, it is constrained by the shared context window and blocking subtask execution, limiting both scalability and adaptability.

\subsection{Parallel Agent Loop}
\label{sec:parallel}

We first introduce the concept of \textit{threads} in a single agent. The original "Think-Act-Observe" trajectory that aims to solve the global task can be viewed as the main thread, denoted $\tau_{\mathrm{main}}$, with context $h_{\mathrm{main}}$. In contrast to classical agent loops, Self-Manager is no longer confined to a single thread, and enables the main thread to dynamically spawn subthreads. Each subthread executes an independent loop, with objectives decoupled from the global task and restricted to the subtask delegated by the main thread, enabling focused tool-integrated reasoning without interference from non-critical information. Upon completion, each subthread returns its subtask result to the main thread, thereby contributing to the global task. The execution flow of Self-Manager is shown in Algorithm \ref{alg:main_thread}.

Concretely, the key design components of Self-Manager's parallel loop include the following:
\begin{itemize}
    \item \textbf{Main thread action space}: During the act phase, the main thread may invoke a tool, such as web retrieval, or manipulate subthreads. Operations on subthreads mainly include creation, termination, and deletion.
    \item \textbf{Main thread observations}: During the observe phase, besides tool feedback or results of thread operation (success/failure), the agent also receives meta information of all existing subthreads. These observations allow the main thread to holistically analyze subthread states and global task progress, thereby determining the action in the next turn. Details of this mechanism are presented in \S \ref{sec:tcb}.
    \item \textbf{Subthreads}: For simplicity, subthreads are not allowed to create nested subthreads. Each subthread executes a standard ReAct loop, as shown in Algorithm \ref{alg:sub_thread}.
\end{itemize}

\newtcolorbox{stepbox}{colback=yellow!20, colframe=black, boxrule=0.5pt, arc=2mm}

\begin{algorithm}[t]
\caption{Self-Manager Agent Execution}
\label{alg:main_thread}
\begin{algorithmic}[1]
\Require Task $x$, agent's policy $\pi_{\theta}$, environment $\mathcal{E}$, maximum turns $T_{\mathrm{max}}$, tools $\mathcal{T}$
\Ensure Final report $\mathcal{A}$

\State Initialize: $h_{\mathrm{main}} \gets \emptyset$, TCB list $\mathcal{B}_0 \gets \emptyset$

\For{$t=1$ to $T_{\mathrm{max}}$}

    \State $a_{t} \sim \pi_{\theta}\left(\cdot \mid x, h_{\mathrm{main}}\right)$ \textcolor{blue}{\Comment{action}}

    \State \textbf{if} $a_t$ is "Finish" \textbf{then}
    \State \quad $\mathcal{A} \gets \pi_{\theta}\left(\cdot \mid x, h_{\mathrm{main}}\right)$; \textbf{break}

    \State \textbf{if} $a_t \in \mathcal{T}$ \textbf{then} $o_t \gets \mathcal{E}\left(a_t\right)$ \textcolor{blue}{\Comment{tool observation}}

    \State \textbf{if} $a_t$ is "Create Threads" \textbf{then}
    \State \hspace{1.0em} \textbf{for} thread $\varsigma \sim a_t$ \textbf{do}
    \State \hspace{1.0em} \hspace{1.0em} \texttt{create\_async\_task}$(\varsigma)$

    \State \hspace{1.0em} \hspace{1.0em} $\mathcal{B}_{t-1} \gets \mathcal{B}_{t-1} \cup \{\varsigma\}$

    \State \textbf{if} $a_t$ is "Kill" \textbf{then} stop $\varsigma \sim a_t$

    \State \textbf{if} $a_t$ is "Del" \textbf{then} $\mathcal{B}_t \gets \mathcal{B}_{t-1} \setminus \{\varsigma \sim a_t\}$

    \State $ o_t \gets o_t \oplus \mathcal{B}_t, h_{\mathrm{main}} \gets (a_t,o_t)$ \textcolor{blue}{\Comment{TCB observation}}

\EndFor

\State \Return $\mathcal{A}$
\end{algorithmic}
\end{algorithm}

\begin{algorithm}[htbp]
\caption{\texttt{create\_async\_task}$(\varsigma)$}
\label{alg:sub_thread}
\begin{algorithmic}[1]
\Require Goal $x'$, available tools $\mathcal{T}'$, prefix context $h'$
\Ensure Answer $\mathcal{A}'$

\State Initialize subthread: $h_{\mathrm{sub}} \gets \{h'\}$
\For{$t=1$ to $T_{\mathrm{max}}$}

    \State $a_{t}' \sim \pi_{\theta}\left(\cdot \mid x', h_{\mathrm{sub}}\right)$ \textcolor{blue}{\Comment{action}}
    \State \textbf{if} $a_t'$ is "Finish" \textbf{then}
    \State \hspace{1em} $\mathcal{A}' \gets \pi_{\theta}\left(\cdot \mid x', h_{\mathrm{sub}}\right)$;\textbf{break}
    \State \textbf{if} $a_t' \in \mathcal{T}' $ \textbf{then}
    \State \hspace{1em} $o_t' \gets \mathcal{E}\left(a_t'\right)$ \textcolor{blue}{\Comment{observation}}
    \State $h_{\mathrm{sub}} \gets (a_t', o_t')$

\EndFor

\State \textbf{return} $\mathcal{A}'$

\end{algorithmic}
\end{algorithm}

Based on the above design, Self-Manager enables several key features of a parallel agent loop:
\begin{itemize}
    \item \textbf{Asynchrony}. Once a subthread is created, the main thread is not blocked. Instead, all threads run asynchronously in parallel. Similar to thread switching in operating systems, a thread only needs to await when its loop requests the base model to generate an action, which constitutes the sole blocking event. During this time, context switching between threads occurs naturally.
    \item \textbf{Concurrency}. The main thread can create multiple subthreads that execute asynchronously, allowing different aspects of a complex task to be explored in parallel.
    \item \textbf{Context Isolation}. Each thread maintains its own dedicated context window, preventing interference between concurrently executing tasks. During initialization, a subthread receives information such as the subtask description provided by the main thread. Upon completion, it returns the subtask execution results to the main thread’s context.
\end{itemize}

Thus, the trajectories and contexts encompassed by Self-Manager can be summarized as follows:
\begin{align}
    &\left( \tau_{\mathrm{main}} = \bigl( (a_t,o_t)_{t=1}^{T},\, \mathcal{A} \bigr), h_{\mathrm{main}} \right),\\
    &\left \{ \tau_{\mathrm{sub}}^{(i)} = \bigl( (a_t^{(i)},o_t^{(i)})_{t=1}^{T^{(i)}},\, \mathcal{A}^{(i)} \bigr), h_{\mathrm{sub}}^{(i)} \right \}^N_{i=1}.
\end{align}

\subsection{Agent-as-TCB-Manager}
\label{sec:tcb}

As described in \S \ref{sec:parallel}, the main thread of Self-Manager achieves self-management of subthreads by presenting the state information of all threads during the Observe phase. This design preserves the generalizable Think–Act–Observe loop structure. Specifically, we draw inspiration from operating systems: each subthread is associated with a \textit{Thread Control Block} (TCB), which is a data structure that contains the thread-specific information required for its management. Replacing raw contextual information with TCBs can significantly improve the efficiency of subthread management.

\begin{wrapfigure}{r}{0.54\textwidth}
\centering
\includegraphics[width=0.54\textwidth]{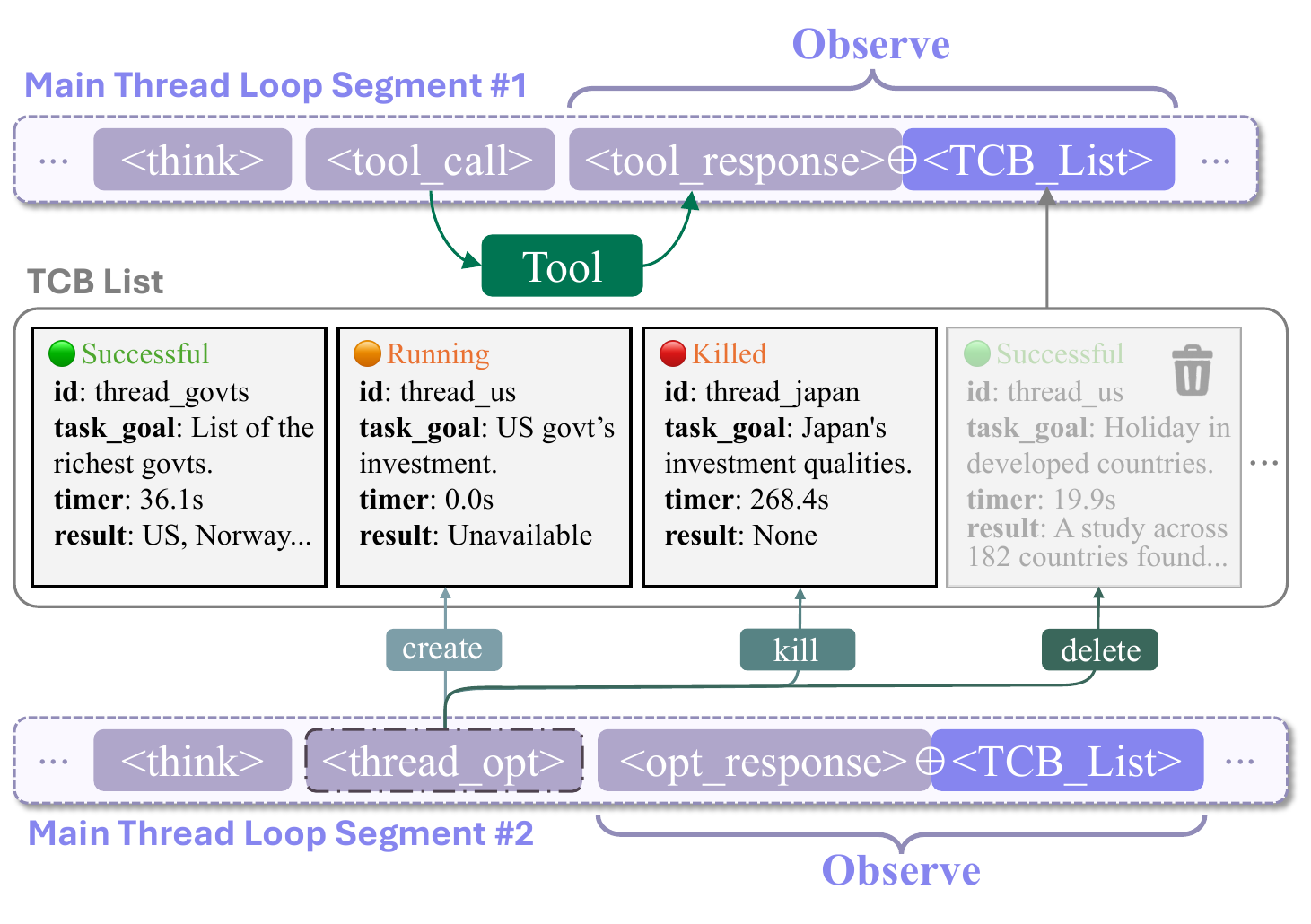}
\caption{An illustration of how Self-Manager manages subthreads via TCBs. In each iteration, the environment response in the Observe phase includes both the feedback from the tool call and the TCB list.}
\label{fig:tcb}
\end{wrapfigure}
As shown in Figure \ref{fig:tcb}, the TCB defines, from the perspective of a thread, the fields that are most critical to the agent loop. First, there is the basic thread metadata, which includes the thread ID, task description, as well as a result field used to present the execution outcome upon completion. In addition to the fields shown in Figure \ref{fig:tcb}, the TCB also contains resource-related information, such as the prefix context inherited from the main thread and the tools assigned to the subthread. Furthermore, it maintains thread state information, such as whether the thread is currently running and the elapsed execution time. Except for the state information, all other fields are specified by the main thread, thereby ensuring flexibility in self-management. More details of the TCB definition and implementation can be found in Appendix \ref{appendix:implementations}.

As depicted in Algorithm \ref{alg:main_thread} and Figure \ref{fig:tcb}, during each observe phase of the main thread in Self-Manager, the latest information from all subthreads's TCBs is accessible. Based on these observations, the main thread can assess the state of each subthread and determine actions in the next turn, such as spawning new threads to handle uncovered subtasks, terminating subthreads that have run excessively long with little benefit, or deleting TCBs that contain no critical information. By integrating the TCB into the agent loop, the main thread can effectively manage threads, thereby continuously advancing the resolution of complex tasks. Compared to parallel execution without autonomous management \citep{schroeder-etal-2025-thread, lian2025threadweaveradaptivethreadingefficient}, the self-managed approach provides enhanced flexibility and task concentration.

Since the states of subthreads evolve in real time, we further reduce context overhead by retaining only the most recent turn of the TCB list within the loop for immediate action decision-making in the next turn. After the next turn of action generation is completed, the previous TCB list is cleared to eliminate interference from outdated state information and to avoid redundancy.

\begin{table*}[t]
\caption{Results on DeepResearch Bench. The best-performing approaches among single-agent baselines and among other baseline categories are highlighted in boldface. Green and red numbers denote the performance gains and drops when using alternative base models compared to Qwen3-30B-A3B, respectively.}
\label{tb:deepresearch_bench}
\resizebox{\linewidth}{!}{
\renewcommand\arraystretch{1.0}

\begin{tabular}{lccccccc}
\toprule
\midrule

\multirow{3}{*}{\textbf{Model}} & \multicolumn{5}{c}{\textbf{RACE}} & \multicolumn{2}{c}{\textbf{FACT}} \\

\cmidrule(lr){2-6} \cmidrule(lr){7-8} 
&  \textbf{Overall}  &  \textbf{Comp.} &  \textbf{Insight}  &  \textbf{Inst.}  &  \textbf{Read.} &  \textbf{C.acc.} & \textbf{Eff.c.}\\

\midrule

\textcolor{gray}{\textit{LLMs w/ Search}} \\
Qwen3-30B-A3B & 41.83 & 39.85 & 40.27 & 44.74 & 44.44 & 60.63 & 8.30 \\
Claude-3.7-Sonnet & 42.29 & 41.45 & 43.78 & 44.53 & 42.23 & 89.51 & 10.39 \\
\hdashline

\textcolor{gray}{\textit{Multi-Agent Workflow}} \\
Langchain-open-DeepResearch & 43.89 & 43.61 & 42.19 & 47.30 & 45.07 & 53.52 & 31.17 \\
AI-Q NVIDIA Research Assistant & 43.69 & 42.07 & 42.76 & 46.15 & 45.14 & - & - \\
\hdashline

\textcolor{gray}{\textit{Proprietary Agents}} \\
Gemini-2.5-Pro Deep Research & \textbf{48.24} & \textbf{48.24} & \textbf{47.62} & 48.74 & \textbf{48.97} & 85.71 & 27.0 \\
OpenAI-DeepResearch & 47.20 & 47.07 & 45.84 & \textbf{48.98} & 47.64 & 65.30 & 34.15 \\
\midrule

\textcolor{gray}{\textit{Single Agent}} \\
ReAct (Qwen3-30B-A3B) & 40.51 & 39.15 & 37.71 & 43.98 & 43.40 & 21.69 & 1.36 \\
ReSum (Qwen3-30B-A3B) & 42.77 & 41.34 & 40.09 & 46.82 & 45.15 & 32.72 & 2.33 \\
FoldAgent (Qwen3-30B-A3B) & 41.79 & 40.04 & 38.95 & 46.22 & 43.97 & 19.61 & 1.56 \\
\rowcolor{lightpink}
Self-Manager (Qwen3-30B-A3B) & \textbf{44.33} & \textbf{43.14} & \textbf{42.28} & \textbf{47.61} & \textbf{45.97} & 25.49 & 2.19 \\
\hdashline
\rowcolor{lightpink}
Self-Manager (Qwen3-8B) & ${\text{43.88}}_{\textcolor{red!50}{\downarrow 0.45}}$ & ${\text{42.30}}_{\textcolor{red!50}{\downarrow 0.84}}$ & ${\text{42.00}}_{\textcolor{red!50}{\downarrow 0.28}}$ & ${\text{47.35}}_{\textcolor{red!50}{\downarrow 0.26}}$ & ${\text{45.78}}_{\textcolor{red!50}{\downarrow 0.19}}$ & 17.80 & 1.86\\
\rowcolor{lightpink}
Self-Manager (Qwen3-30B-A3B-2507) & ${\text{45.47}}_{\textcolor{mygreen}{\uparrow 1.14}}$ & ${\text{43.50}}_{\textcolor{mygreen}{\uparrow 0.36}}$ & ${\text{44.07}}_{\textcolor{mygreen}{\uparrow 1.79}}$ & ${\text{48.21}}_{\textcolor{mygreen}{\uparrow 0.60}}$ & ${\text{48.08}}_{\textcolor{mygreen}{\uparrow 2.11}}$ & 24.80 & 2.51 \\
\rowcolor{lightpink}
Self-Manager (Qwen3-Next-80B-A3B) & ${\text{45.63}}_{\textcolor{mygreen}{\uparrow 1.30}}$ & ${\text{43.53}}_{\textcolor{mygreen}{\uparrow 0.39}}$ & ${\text{44.32}}_{\textcolor{mygreen}{\uparrow 2.04}}$ & ${\text{48.17}}_{\textcolor{mygreen}{\uparrow 0.56}}$ & ${\text{48.14}}_{\textcolor{mygreen}{\uparrow 2.17}}$ & 35.90 & 6.69  \\

\midrule
\bottomrule
\end{tabular}
}
\end{table*}

\section{Experiments}

In this section, we conduct experiments centered on Self-Manager to investigate the following research questions: (1) How well does Self-Manager perform on complex tasks? (2) How do the individual components of Self-Manager affect its overall performance? (3) How effectively does Self-Manager perform in terms of contextual capabilities? (4) What are the efficiency and computational cost of Self-Manager? (5) How well does Self-Manager generalize across different tasks?

\subsection{RQ1: Performance on Complex Tasks}

This section investigates the performance of our proposed Self-Manager on complex tasks, particularly those that require subtask decomposition for effective problem solving.

\paragraph{Setup.} We evaluate our method on DeepResearch Bench \citep{du2025deepresearchbenchcomprehensivebenchmark}, which consists of 100 complex long-form deep research tasks. We use Gemini-2.5-Flash \citep{gemini-2.5-flash} as the LLM-as-Judge in evaluation. We compare Self-Manager against several categories of baselines: (1) \textit{search-augmented LLMs}; (2) \textit{single agent}, referring to agent loops such as ReAct \citep{yao2022react}, ReSum \citep{wu2025resumunlockinglonghorizonsearch}, and FoldAgent \citep{sun2025scaling}. Single agents are configured with a maximum context window of 128k tokens, using Serper\footnote{\url{https://serper.dev/}} as the search tool and Jina Reader\footnote{\url{https://jina.ai/reader/}} as the browsing tool; (3) \textit{multi-agent workflows}, which consist of fixed, predefined pipelines of multiple agents, including LangChain-Open-DeepResearch \citep{langchain_open_deepresearch} and AI-Q Research Assistant \citep{aiq}; (4) \textit{proprietary deep research agents}, namely state-of-the-art deep research systems, including Gemini Deep Research \citep{gemini_deep_research} and OpenAI DeepResearch \citep{openai2025deepresearch}.

\paragraph{Results.} As shown in Table \ref{tb:deepresearch_bench}, Self-Manager outperforms all single-agent architectures and surpasses workflow-based multi-agent systems, while further narrowing the performance gap with proprietary deep research agents. These findings highlight the advantage of Self-Manager’s parallel agent loop in handling difficult long-form deep research tasks. In addition, we assess Self-Manager with base models of varying sizes and training strategies. We observe that stronger model capabilities correspond to better performance, implying that Self-Manager does not limit the expressive power of the base model and adheres to scaling rules.

\begin{figure*}
    \begin{minipage}[b]{0.30\textwidth}
    \centering
    \includegraphics[width=1.0\textwidth]{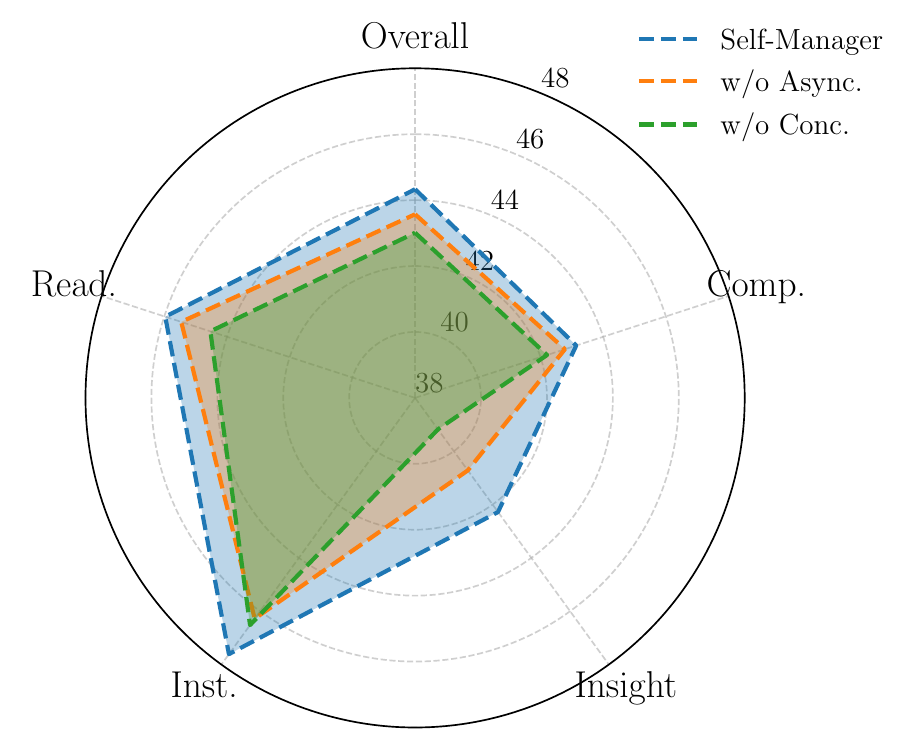}
    \caption{Ablation results on the asynchronous and concurrent capabilities of Self-Manager, showing performance degradation.}
    \label{fig:ablation}
    \end{minipage}
    \hfill
    \begin{minipage}[b]{0.67\textwidth}
    \centering
    \includegraphics[width=1.0\textwidth]{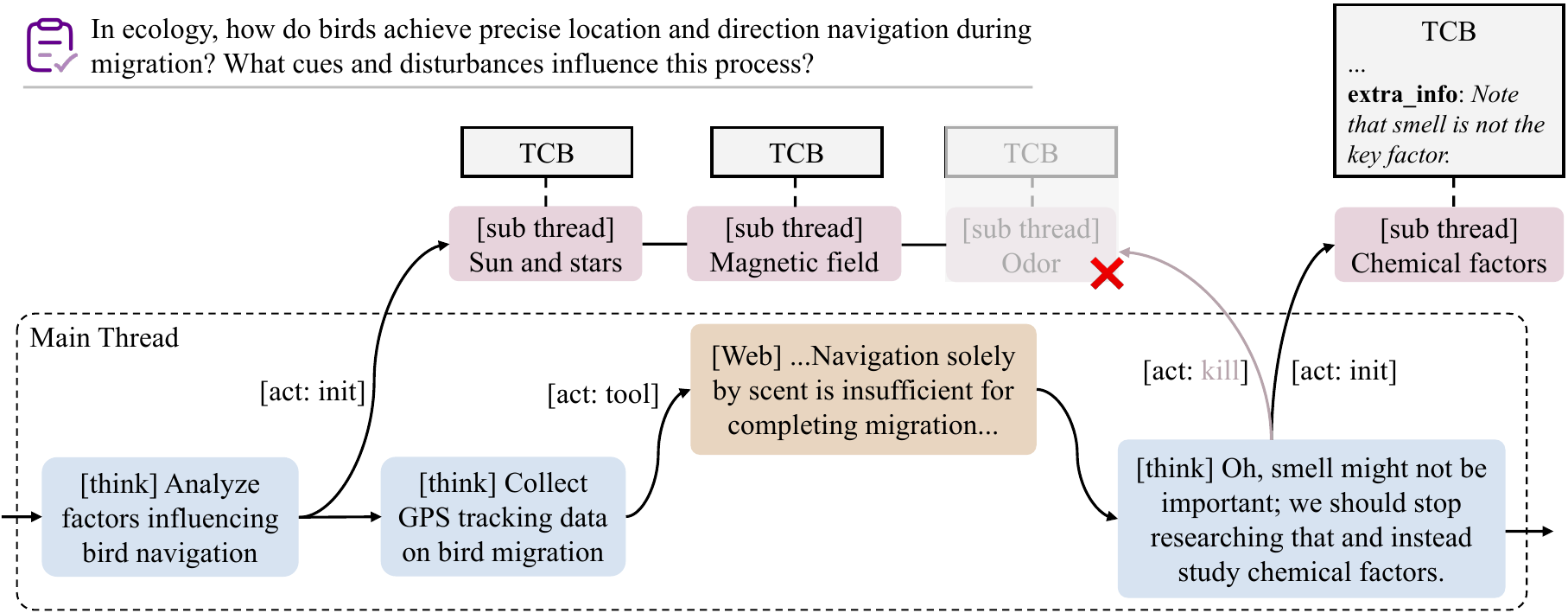}
    \caption{A Case in Asynchronous Parallelism of Self-Manager. It shows that the main thread determines that a running subthread is no longer necessary for the overall task and then terminates it. This self-management mechanism enables early stopping, making execution more adaptive and efficient.}
    \label{fig:ablation_case}
    \end{minipage}
\end{figure*}

\subsection{RQ2: Ablation Studies}

This section presents a series of ablation studies centered on the design of Self-Manager, with the goal of validating the necessity and effectiveness of its core components.

\paragraph{Setup.} We conduct ablations at two levels. The first level targets the agent loop design of Self-Manager, with the following baselines: (1) \textit{without asynchrony}, where the main thread becomes fully blocked whenever subthreads are running, and the agent loop resumes only after all subthreads have completed; (2) \textit{without concurrency}, where at most one thread is allowed to run and only one subthread can be created at any time. The second level focuses on the design of the TCB, with the following baselines: (1) \textit{without assigned\_tool}, where the TCB does not specify the set of tools available to the current thread, and the thread is instead allowed to access all tools available to the main thread; (2) \textit{without extra\_info}, where the main thread does not provide any additional information to subthreads; (3) \textit{prefix\_context variants}, which have four designs: providing the main thread's full, no, or summarized prefix context, or allowing the main thread to freely select one of the three choices listed above.

\begin{wrapfigure}{r}{0.55\textwidth}
\centering
\includegraphics[width=0.55\textwidth]{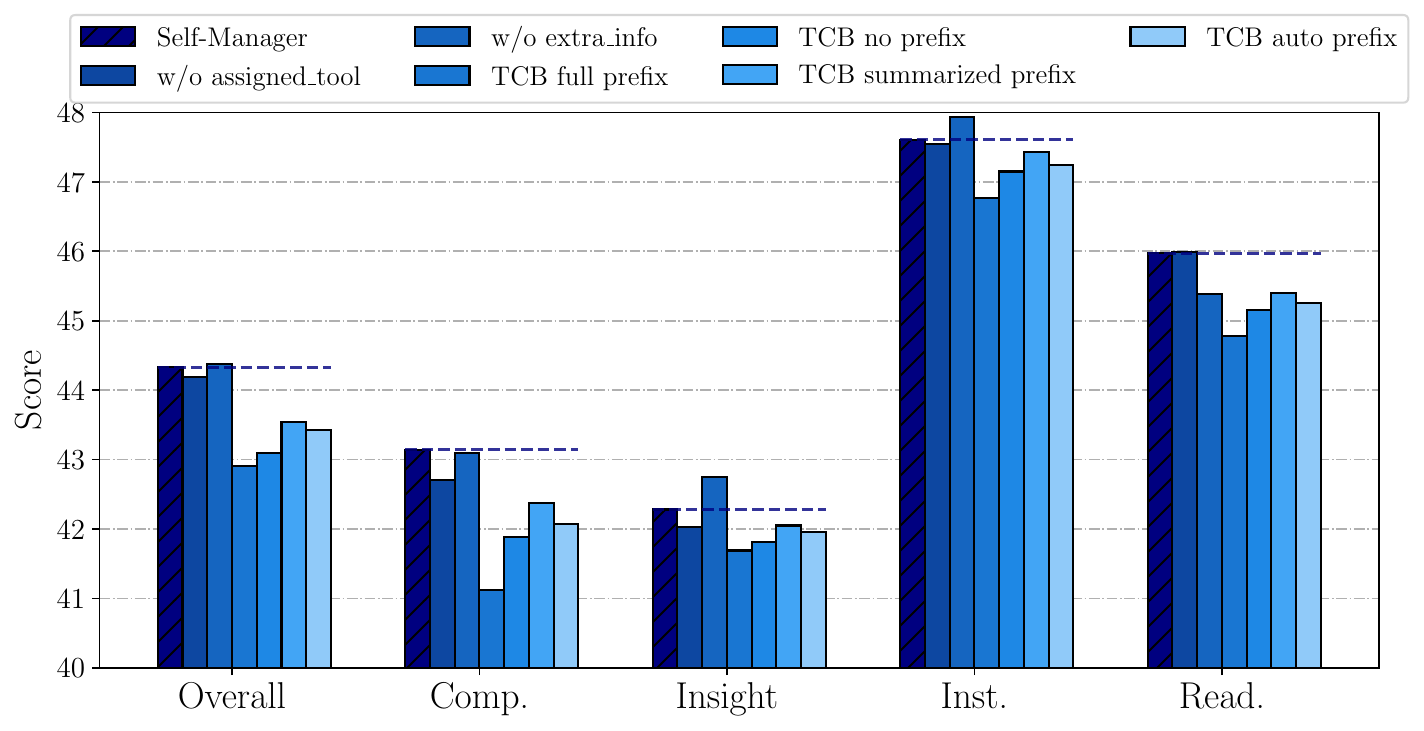}
\caption{Ablation study on the design of the Thread Control Block (TCB).}
\label{fig:ablation_tcb}
\end{wrapfigure}
\paragraph{Results.} The results in Figures \ref{fig:ablation} and \ref{fig:ablation_tcb} show that Self-Manager consistently outperforms all ablation baselines across multiple metrics on DeepResearch Bench, demonstrating the effectiveness of its asynchronous execution, concurrency, and TCB-based management design. While some variants achieve performance close to that of the original Self-Manager, this does not imply design redundancy, which can be verified by the following cases.

\paragraph{Case Study.} As illustrated by the case in Figure \ref{fig:ablation_case}, when subthreads are running, the asynchronous main thread continues to perform work that makes tangible contributions to the overall task. Specifically, the main thread identifies that the subtask being executed by a running subthread is unnecessary and thus performs an early stop, preventing wasted computational resources. Moreover, the \textit{extra\_info} field can occasionally encode note-like hints that better guide subthreads in completing their assigned subtasks. Therefore, although no absolute performance gain, we retain these design choices in order to improve the adaptiveness.

\subsection{RQ3: Contextual Capabilities}

\paragraph{Information Retention.} For loops in which the context may undergo changes, it is necessary to consider the information loss rate at each context transition. To quantify information loss, we adopt the LLM-as-Judge to dynamically maintain a list of valid information units. At each context change, we compare the information sets before and after the transition as follows:
\begin{equation}
\label{eq:info_retention}
    \mathcal{L}_{\mathrm{info} } = 1 - \frac{\left | \mathcal{I}_j \cap \tilde{\mathcal{I}_j }   \right | }{\left | \mathcal{I}_j  \right | },
\end{equation}
where $\mathcal{I}_j$ and $\tilde{\mathcal{I}_j }$ denote the sets of information units before and after the $j$-th context change, respectively. More details are provided in Appendix \ref{appendix:experiment}. Experimental results in Table \ref{tb:info_loss} show that Self-Manager achieves a lower loss rate through context isolation, while the main thread maintains a consistently smaller accumulated context length.

\begin{figure*}[t]
    \begin{minipage}[b]{0.30\textwidth}
    \centering
    \includegraphics[width=1.0\textwidth]{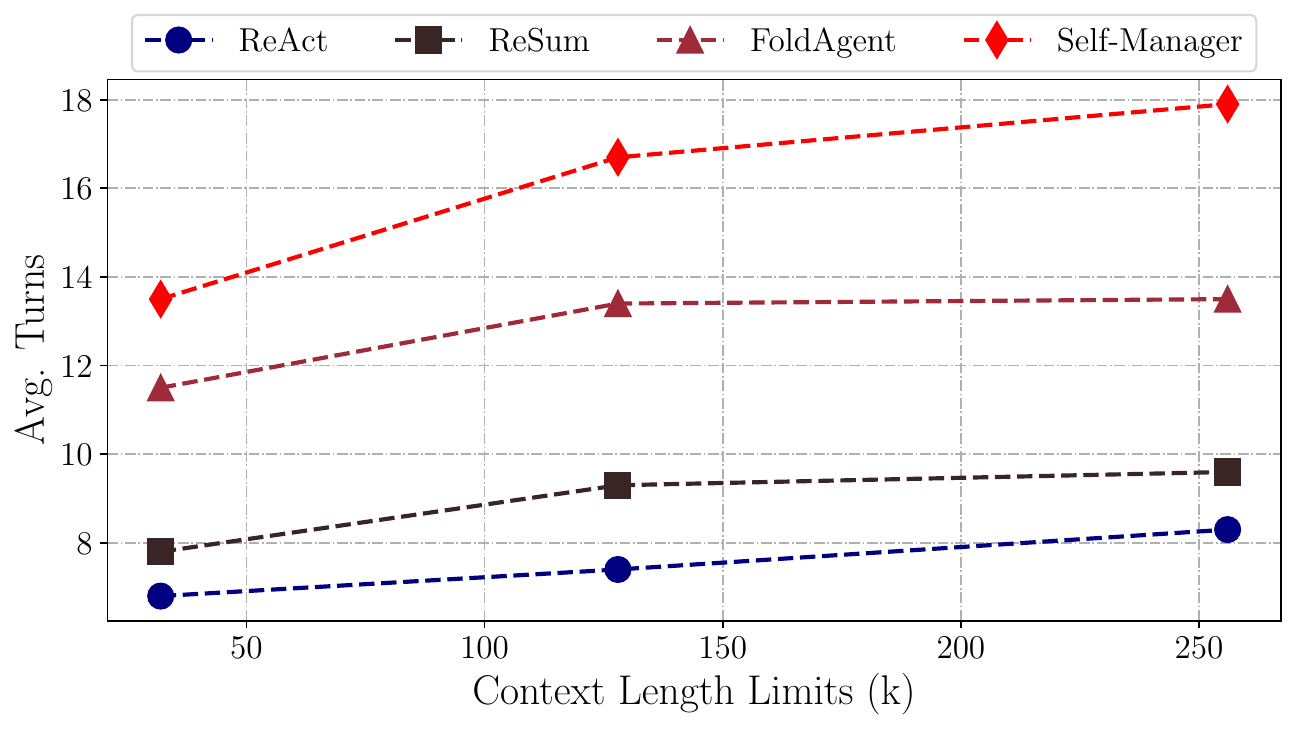}
    \caption{Comparison of the average execution turns across different agent loop architectures under varying context window length limits.}
    \label{fig:long_horizon}
    \end{minipage}
    \hfill
    \begin{minipage}[b]{0.67\textwidth}
    \centering
    \includegraphics[width=1.0\textwidth]{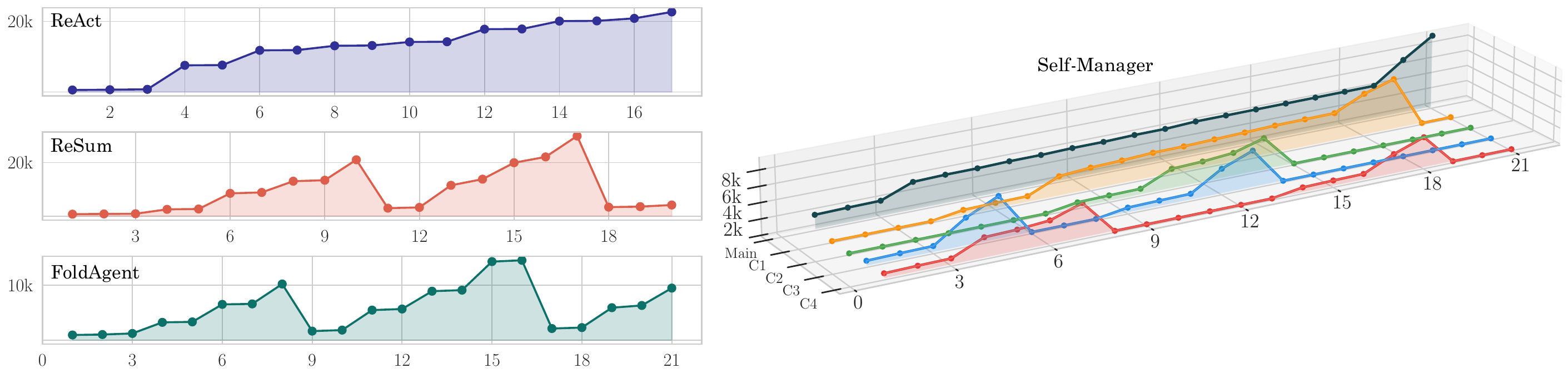}
    \caption{Examples of context accumulation across agent loops. The x-axis denotes the turns, while the y-axis denotes the context length. In Self-Manager, "Main" represents the main thread's context, whereas "C" entries correspond to the slots of subthreads. In this example, up to four subthreads run concurrently.}
    \label{fig:context_lines}
    \end{minipage}
\end{figure*}

\begin{wraptable}{r}{0.55\textwidth}
\centering
\caption{Context capability assessment. C.C. denotes the count of context changes, including summarization, context folding, and subthread creation. Avg. PL refers to the average peak context length, and Avg. L denotes the average context length after execution.}
\label{tb:info_loss}
\resizebox{0.55\textwidth}{!}{
\renewcommand\arraystretch{1.1}
\begin{tabular}{lcccc}
\toprule
\midrule
\textbf{Model} & \textbf{$\mathcal{L}_{\mathrm{info} } \downarrow$} & \textbf{C.C.} & \textbf{Avg. PL} & \textbf{Avg. L} \\ 
\midrule

ReAct & - & - & 13.1k & 13.1k \\
ReSum & 25.23\% & 2.7 & 14.2k & 15.7k \\
FoldAgent & 19.24\% & 5.6 & 9.5k & 9.5k \\
\hdashline

\rowcolor{lightpink}
Self-Manager & \textbf{11.53\%} & 7.0 & 10.2k & 10.2k \\

\midrule
\bottomrule
\end{tabular}
}
\end{wraptable}
\paragraph{Long-Horizon.} We measure the average count of execution turns under different context window length constraints to assess how well the agent loop supports long-horizon execution. The results in Figure \ref{fig:long_horizon} show that the main thread of Self-Manager consistently runs for more rounds, which can stem from its subtask delegation mechanism and design that allows the main thread to focus solely on the overall task, resulting in slower context accumulation and more long-horizon planning.

\paragraph{Case Study.} Figure \ref{fig:context_lines} illustrates an example of how the context length evolves over time. Vanilla ReAct exhibits linear growth and is prone to exceeding the limit. ReSum and FoldAgent achieve non-monotonic accumulation through compression and folding. In contrast, Self-Manager leverages concurrent subthreads and context isolation, leading to a reduced context accumulation rate in the main thread and exhibiting long-horizon capacity.

\subsection{RQ4: Efficiency and Cost}

This section examines the runtime efficiency and cost of Self-Manager.

\paragraph{Setup.} To fairly measure execution time and costs, we adopt a trajectory-based accounting method that abstracts away external factors such as network fluctuations. Specifically, each agent’s think and act steps are quantified based on the lengths of prefix tokens, generated tokens, and the number of tool invocations, as formalized below:
\begin{align}
T_{\mathrm{total} }&=\sum T_{\mathrm{pre} } + \sum T_{\mathrm{gen} } +  \sum_{\mathrm{tool}_i \in \mathcal{T}} \sum T_{\mathrm{tool}_i }\\
C_{\mathrm{total} }&=\sum C_{\mathrm{pre} } + \sum C_{\mathrm{gen} } +  \sum_{\mathrm{tool}_i \in \mathcal{T}} \sum C_{\mathrm{tool}_i }
\end{align}
Calculating details are provided in Appendix \ref{appendix:experiment}.

\paragraph{Results.} As indicated in Table \ref{tb:efficiency}, relative to FoldAgent, Self-Manager achieves better task performance while demonstrating controllable runtime and cost. Although the overall overhead of Self-Manager is higher, superior performance on long-form deep research tasks generally comes at the expense of increased tool invocations, which we consider acceptable.

\subsection{RQ5: Generalization Ability}

This section investigates the generalization of Self-Manager across other types of tasks.

\paragraph{Setup.} Although Self-Manager is inherently designed for complex tasks requiring subtask decomposition, its general architecture can be applied to a wide range of scenarios. To evaluate this, we select BrowseComp-Plus \citep{chen2025browsecomp}, a benchmark requiring multi-hop deep search. 

\paragraph{Results.} The experiments show that Self-Manager achieves the best performance among single agents on this benchmark. This can be attributed to the parallel agent loop, which not only allows fine-grained management of context to reduce the difficulty of complex tasks but also adapts to the task complexity by controlling whether subthreads are created or if the main thread alone suffices. These findings indicate that Self-Manager exhibits generalization and adaptability across complex deep research tasks.

\begin{minipage}[c]{0.48\textwidth}
\centering
\captionof{table}{Comparison of efficiency and cost between Self-Manager and FoldAgent. RN denotes the number of subtask result returns.}
\resizebox{\linewidth}{!}{
\renewcommand\arraystretch{1.1}

\begin{tabular}{lcccc}
\toprule
\midrule

\textbf{Model} & \textbf{Time} (s) & \textbf{Cost} (\$) & \textbf{Tool Calls} & \textbf{RN} \\
\midrule

FoldAgent & 117.27 & 0.05 & 18.25 & 5.6 \\
\hdashline

\rowcolor{lightpink}
Self-Manager & 129.62 & 0.09 & 20.30 & 7.0 \\
\quad \textit{w/o} Async & 193.08 & 0.07 & 21.63 & 7.2 \\
\quad \textit{w/o} Conc & 144.55 & 0.07 & 19.37 & 6.6 \\

\midrule
\bottomrule
\end{tabular}
}
\label{tb:efficiency}
\end{minipage}
\hfill
\begin{minipage}[c]{0.5\textwidth}
\centering
\captionof{table}{Results on BrowseComp-Plus.}
\resizebox{\linewidth}{!}{
\renewcommand\arraystretch{1.1}

\begin{tabular}{lccc}
\toprule
\midrule

\textbf{Model} & \textbf{Acc. (Pass@1)} & \textbf{Recall} & \textbf{Tool Calls} \\
\midrule

\multicolumn{4}{l}{\textcolor{gray}{\textit{LLMs w/ Search}}} \\
Qwen3-30B-A3B & 5.32\% & 6.70\% & 1.00 \\
\hdashline

\multicolumn{4}{l}{\textcolor{gray}{\textit{Single Agent} (Qwen3-30B-A3B)}} \\
ReAct & 28.47\% & 37.33\% & 5.79 \\
ReSum & 31.24\% & 38.90\% & 6.32 \\
FoldAgent & 35.65\% & 43.04\% & 12.15 \\
\rowcolor{lightpink}
Self-Manager & \textbf{36.17\%} & \textbf{45.39\%} & 18.45 \\

\midrule
\bottomrule
\end{tabular}
}
\label{tb:browsecomp}
\end{minipage}

\section{Related Work}

\paragraph{Web Agents.} While Retrieval-Augmented Generation \citep[RAG;][]{lewis2021retrievalaugmentedgenerationknowledgeintensivenlp, gao2024retrievalaugmentedgenerationlargelanguage, xu2025training} improves the factuality of Large Language Models \citep[LLMs;][]{naveed2025comprehensive, mei2024not, mei2024slang, xu2024aliice}, its fixed and passive retrieval paradigms remain limited. In contrast, web agents \citep{wu2025webdancerautonomousinformationseeking, li2025websailornavigatingsuperhumanreasoning} adopt an agent-centric architecture to perform autonomous and adaptive multi-turn information seeking and synthesis. \citet{li2025search, li2025webthinker} integrate reasoning with web tool invocation, leading to notable performance improvements. \citet{jin2025searchr1trainingllmsreason, song2025r1, gao2025turnsunlockinglonghorizonagentic} further enhance web agents through reinforcement learning. \citet{zhang2025processvsoutcomereward, zhang2025letslearningthinkandsearchprocessandoutcome} investigate process-level rewards to increase reward density, thereby providing more fine-grained guidance. Currently, most web agents primarily address multi-hop QA tasks, such as HotpotQA \citep{yang2018hotpotqa} and MuSiQue \citep{trivedi2022musique}.

\paragraph{Deep Research.} Compared to traditional QA, deep research is more complex and must contend with information at web scale. \citet{zheng2025deepresearcherscalingdeepresearch, mialon2023gaiabenchmarkgeneralai} trains and evaluates deep research agents on real-world web corpora. Existing deep research tasks can be broadly categorized into two types. The first type is short-form, where the target answer is typically a short entity, in order to assess the information-seeking capacity by issuing queries with numerous constraints \citep{wei2025browsecompsimplechallengingbenchmark, chen2025browsecomp, chen2025xbenchtrackingagentsproductivity}. The second type is long-form, where the desired output is usually a report \citep{du2025deepresearchbenchcomprehensivebenchmark, xu2025ravine, wang2025liveresearchbenchlivebenchmarkusercentric}. Deep research tasks generally require dedicated multi-agent workflows to be effectively addressed \citep{prabhakar2025enterprisedeepresearchsteerable, hong2025multi}.

\paragraph{Context Management.} As agents handle increasingly complex tasks, the accumulated context becomes increasingly lengthy, which can dilute salient information and hinder effective decision-making \citep{liu-etal-2024-lost, zhang2025agenticcontextengineeringevolving}. Consequently, context management has become a critical challenge \citep{mei2025surveycontextengineeringlarge}. Some work focuses on long-text modeling \citep{mei2026gateddifferentiableworkingmemory}, safety \citep{mei2024hiddenguard}, and related issues, while the context of agents has also received significant attention. \citet{wu2025resumunlockinglonghorizonsearch, nguyen2025sfrdeepresearcheffectivereinforcementlearning} adopt summary-based compression strategies to mitigate linear context growth. \citet{ye2025agentfoldlonghorizonwebagents, sun2025scaling} perform context folding at the granularity of subtasks. \citet{zhang2025agenticcontextengineeringevolving, chen2025iterresearchrethinkinglonghorizonagents} replace conversational histories with structured context representations that explicitly preserve key information. Another line of work treats memory as an extension of context \citep{li2025memos, chhikara2025mem0buildingproductionreadyai, hu2025evaluating}, enabling long-term memory retrieval and management. Our proposed Self-Manager aims to further optimize single agents through parallelized execution, addressing limitations of prior solutions such as context interference, information loss, and synchronous execution.

\section{Conclusion}

This work proposes Self-Manager, a novel parallel agent loop architecture based on a thread-oriented design. The main thread decomposes a complex task into multiple subtasks and dispatches them to subthreads for execution, thereby addressing key limitations of prior approaches, including context accumulation explosion, information loss, inter-task interference, and blocking execution. More importantly, Self-Manager introduces a Thread Control Block (TCB) mechanism to enable self-management of subthreads, resulting in more efficient and adaptive asynchronous parallel execution. We evaluate the effectiveness of Self-Manager on the DeepResearch Bench, and further investigate the necessity of its modular design, as well as its advantages in terms of contextual capacity, efficiency, and generalization. We hope that Self-Manager can serve as an insightful agent architecture for future research on parallel agents.

\newpage

\bibliography{custom}

\appendix
\clearpage

\section{Introduction of Classical Agent Loops}

Some prior work allows the agent to summarize the context either during the act step or when approaching the context-window limit \citep{wu2025resumunlockinglonghorizonsearch}, leading to the modified trajectory:
\begin{equation}
    \left(\dots, o_{i-1}, a_{i}=\mathrm{sum} \right) \rightarrow\left(h^{\prime}_{i-1}, a_{i+1}, o_{i+1}, \ldots\right),
\end{equation}
where $h^{\prime}_{i-1}$ represents the compressed historical context before turn $i$. Yet this introduces a fundamental difficulty: summarization performs lossy compression, and selecting which information to preserve is inherently challenging.

To alleviate this, FoldAgent \citep{sun2025scaling} proposes to fold the context along subtask boundaries, thereby reducing linear context growth at the granularity of clearly useful information. Formally:
\begin{equation}
    ({\color{red} a_{i}}, \underbrace{{\color{red} o_{i}}, \dots, {\color{purple} a_{j}}}_{\mathrm{subtask} },  {\color{purple} o_{j}}) \xrightarrow{\mathrm{Folding}} ({\color{red} a_{i}}, {\color{purple} o_{j}}, a_{j+1}, o_{j+1} \dots).
\end{equation}

However, as the subtask and the global trajectory still share the same context window, information interference persists, and subtask execution remains synchronous and blocking, which limits efficiency. To address the above limitations, we present Self-Manager, a novel parallel agent loop.

\section{Implementation Details}
\label{appendix:implementations}

In this section, we provide additional details on the Self-Manager method and its implementation.

\subsection{Prompt Templates}

First, the main thread uses a system prompt template that supports not only basic tool invocation but also actions such as thread creation and management, as seen below. Note that the tool definition section of the following prompt template has been relocated to Table \ref{tb:tool_definition}.

\newcommand{\monospace}[1]{{\fontfamily{qcr}\selectfont #1}}

\begin{tcolorbox}[breakable, colback=gray!10, colframe=black, title=\textbf{Prompt Template for Main Thread}]
You are an advanced agent capable of creating subthreads, specifically designed to perform deep research tasks. As the main thread, you operate based on the standard ReAct Loop: Think-Act-Observe. During the Act phase, you may call tools or create subthreads to complete the subtasks you assign. You excel at constructing and managing subthreads, enabling them to focus on researching specific subtopics or to carry out detailed writing for particular sections of the final report.\\\\
Task Description:\\
Given a user's question, your task is to think iteratively based on the question, search for and integrate external web information, and ultimately produce a comprehensive, in-depth, and well-structured long-form report. When you have gathered sufficient information and are ready to provide the definitive long-form report, you must enclose the entire report within <answer></answer> tags.\\\\
Available Tools:\\
You may call a tool function in each turn to assist with the user query. You are provided with function signatures within <tools></tools> XML tags:\\
<tools>...</tools>\\
For each function call, return a json object with function name and arguments within <tool\_call></tool\_call> XML tags:\\
<tool\_call>\\
\{"name": <function-name>, "arguments": <args-json-object>\}\\
</tool\_call>\\\\
Observe:\\
- After you invoke a tool, the Observe phase will provide you with the tool-invocation details, including the returned result and any potential errors, which are enclosed within the <tool\_response></tool\_response> XML tags.\\
- In addition, you can also see the list of TCBs (Thread Control Blocks), each corresponding to the current state of a subthread created by the main thread. Each TCB includes: Thread ID, Target, Status (Running, Success, Failed, Killed), Allowed Tools, Assigned Context, Runtime, and Result (available only after the thread has completed execution).\\
- With the observation information, you can continue to determine your next Action in the loop.\\\\
Report Requirements:\\
1. Your report must be in Markdown format, well-structured, and fluent.\\
2. Your report must align with the intent of the user's question, and can comprehensively address the question.\\
3. Your report should not simply be a list of arguments. For each point of your report, it's not enough to just state the argument --- you need to provide in-depth analysis, causal reasoning, impacts and trends analysis, solutions, and so on. In short, make the description more detailed and substantial.\\
4. Your report should include Markdown-formatted citations for all referenced web sources. For example: ([title](url)).\\
5. The language of your report should be consistent with the language of the user's questions.\\
6. You must enclose the entire report within <answer></answer> tags.\\\\
User's task: \textcolor{blue}{\monospace{\{user\_task\}}}.
\end{tcolorbox}

Specifically, the definitions of all actions allowed in the main thread loop are provided in Table \ref{tb:tool_definition}, including the basic retrieval tools used for deep research as well as operations for managing subthreads.

In contrast, the system prompt template for subthreads does not need to consider thread-related actions. Instead, it focuses on implementing a basic loop that integrates tool-integrated reasoning and ReAct's "Think--Act-Observe" paradigm.

\begin{tcolorbox}[breakable, colback=blue!10, colframe=blue, title=\textbf{Prompt Template for Subthread}]
You are a deep research assistant. You can operate based on the standard ReAct Loop: Think-Act-Observe. You are responsible for completing the task assigned to you. This task is typically part of a deep-research task, but you should remain fully focused on the specific task given to you and disregard anything outside its scope. You must ensure that your output strictly satisfies all requirements of the task.\\\\
Requirements:\\
1. You are allowed to call tools, but only the tools explicitly specified by the user. You must not call any tools outside of the ones provided.\\
2. You may perform multiple iterations to pursue deeper investigation and achieve higher-quality results.\\
3. Your submitted results are recommended to be in Markdown format. When referencing web information, include Markdown-formatted citations for all sources. For example: ([title](url)).\\
4. When you determine that the task can be considered complete, you must enclose the entire submission within <answer></answer> tags.\\
5. The language of your submission text should be consistent with the language of the task.\\
6. Your report should not simply be a list of items; it should provide analysis and causal support for each point or information, and offer solutions when necessary.\\\\
Available Tools:\\
You may call a tool function in each turn to assist with the user query. You are provided with function signatures within <tools></tools> XML tags:
<tools>...</tools>\\
Note that the tools listed above are the complete set; you can only call the tools explicitly specified by the user. For each function call, return a json object with function name and arguments within <tool\_call></tool\_call> XML tags:\\
<tool\_call>\\
\{"name": <function-name>, "arguments": <args-json-object>\}\\
</tool\_call>\\\\
Your task: \textcolor{blue}{\monospace{\{goal\}}}.\\
Your allowed tools: \textcolor{blue}{\monospace{\{allowed\_tools\}}}.\\
Your assigned context: \textcolor{blue}{\monospace{\{assigned\_context\}}}.\\
Extra info: \textcolor{blue}{\monospace{\{extra\_info\}}}.\\
\end{tcolorbox}

In addition, we incorporate runtime error protection into the implementation of Self-Manager. Specifically, when the execution is about to exceed the context window limit, we invoke an external LLM to compress the existing contextual content and replace the original context history with the compressed version. We use Gemini-2.5-Flash as the summarization model, with the prompt shown below.

\begin{tcolorbox}[breakable, colback=orange!10, colframe=orange, title=\textbf{Prompt Template for Context Compression}]
You are an expert at analyzing conversation history and extracting relevant information. Your task is to thoroughly evaluate the conversation history and current question to provide a comprehensive summary that will help answer the question.\\\\
Task Guidelines\\
1. Information Analysis:\\
   - Carefully analyze the conversation history to identify truly useful information.\\
   - Focus on information that directly contributes to answering the question.\\
   - Do NOT make assumptions, guesses, or inferences beyond what is explicitly stated in the conversation.\\
   - If information is missing or unclear, do NOT include it in your summary.\\\\
2. Summary Requirements:\\
   - Extract only the most relevant information that is explicitly present in the conversation.\\
   - Synthesize information from multiple exchanges when relevant.\\
   - Only include information that is certain and clearly stated in the conversation.\\
   - Do NOT output or mention any information that is uncertain, insufficient, or cannot be confirmed from the conversation.\\
3. Output Format: Your response should be structured as follows:\\\\
<summary>\\
- Essential Information: [Organize the relevant and certain information from the conversation history that helps address the question.]\\
</summary>\\\\
Strictly avoid fabricating, inferring, or exaggerating any information not present in the conversation. Only output information that is certain and explicitly stated.\\\\
Question: \textcolor{blue}{\monospace{\{question\}}}\\
Conversation History: \textcolor{blue}{\monospace{\{recent\_history\_messages\}}}\\
Please generate a comprehensive and useful summary. Note that you are not permitted to invoke tools during this process. Use the language of the question to generate the summary.
\end{tcolorbox}

\begin{table*}[h]
\center
\caption{The action list of the main thread in Self-Manager. It encompasses tools for deep research tasks as well as operations for subthread management.}
\label{tb:tool_definition}
\small
\begin{tabular}{p{0.10\linewidth}  p{0.80\linewidth}}
\toprule
\textbf{Action} & \textbf{Definition} \\

\hline

\multirow{7}{*}{search} & \multirow{7}{\linewidth}{\texttt{\{"type": "function", "function": \{"name": "search", "description": "Perform Google web searches then returns a string of the top search results. Accepts multiple queries.", "parameters": \{"type": "object", "properties": \{"query": \{"type": "array", "items": \{"type": "string", "description": "The search query."\}, "minItems": 1, "description": "The list of search queries."\}\}, "required": ["query"]\}\}\}}} \\
& \\
& \\
& \\
& \\
& \\
& \\
\hline

\multirow{8}{*}{visit} & \multirow{8}{\linewidth}{\texttt{\{"type": "function", "function": \{"name": "visit", "description": "Visit webpage(s) and return the summary of the content.", "parameters": \{"type": "object", "properties": \{"url": \{"type": "array", "items": \{"type": "string"\}, "description": "The URL(s) of the webpage(s) to visit. Can be a single URL or an array of URLs."\}, "goal": \{"type": "string", "description": "The specific information goal for visiting webpage(s)."\}\}, "required": ["url", "goal"]\}\}\}}} \\
& \\
& \\
& \\
& \\
& \\
& \\
& \\
\hline

\multirow{16}{*}{branch} & \multirow{16}{\linewidth}{\texttt{\{"type": "function", "function": \{"name": "branch", "description": "Create a subthread to perform a specific task.", "parameters": \{"type": "object", "properties": \{"id": \{"type": "string", "description": "The ID of the subthread. You can generate it freely according to your own habits."\}, "target": \{"type": "string", "description": "The target of the subthread. It must be specific and useful to the user's task."\}, "allowed\_tools": \{"type": "array", "items": \{"type": "string", "description": "The name of an allowed tool for the subthread."\}, "minItems": 1, "description": "The list of allowed tools for the subthread."\}, "assigned\_context": \{"type": "string", "description": "The history context assigned by main thread for the subthread."\}, "extra\_info": \{"type": "string", "description": "Any extra information that the main thread wants to provide to the child thread."\}\}, "required": ["id", "target", "allowed\_tools", "assigned\_context"]\}\}\}}} \\
& \\
& \\
& \\
& \\
& \\
& \\
& \\
& \\
& \\
& \\
& \\
& \\
& \\
& \\
& \\
\hline

\multirow{7}{*}{sleep} & \multirow{7}{\linewidth}{\texttt{\{"type": "function", "function": \{"name": "sleep", "description": "Sleep for a specified duration when you think the only thing to do is wait for the subthread to complete its task.", "parameters": \{"type": "object", "properties": \{"sleep\_duration": \{"type": "number", "description": "The duration in seconds to sleep. Maximum 60 seconds"\}\}, "required": ["sleep\_duration"]\}\}\}}} \\
& \\
& \\
& \\
& \\
& \\
& \\
\hline

\multirow{6}{*}{kill} & \multirow{6}{\linewidth}{\texttt{\{"type": "function", "function": \{"name": "kill", "description": "Kill a running subthread from the TCB list when you think it is no longer needed.", "parameters": \{"type": "object", "properties": \{"id": \{"type": "string", "description": "The ID of the subthread to kill."\}\}, "required": ["id"]\}\}\}}} \\
& \\
& \\
& \\
& \\
& \\
\hline

\multirow{6}{*}{delete} & \multirow{6}{\linewidth}{\texttt{\{"type": "function", "function": \{"name": "delete", "description": "Delete the information of a finished subthread from the TCB list when you think it is no longer needed.", "parameters": \{"type": "object", "properties": \{"id": \{"type": "string", "description": "The ID of the subthread to delete."\}\}, "required": ["id"]\}\}\}}} \\
& \\
& \\
& \\
& \\
& \\
\hline

\end{tabular}
\end{table*}

\subsection{Definition of TCB}
\label{appendix:tcb}

The design of the Thread Control Block (TCB) is inspired by the TCB in operating systems, which serves as a core data structure for managing threads. In Self-Manager, the TCB functions as a metadata encapsulation structure, providing the main thread with real-time information about subthreads to facilitate their management. Unlike in operating systems, where TCBs need to handle low-level resources such as pointers, TCBs in Self-Manager retain only the most essential fields for simplicity and efficiency, including:
\begin{itemize}
    \item \textit{id}: A unique identifier for a subthread, such as "Thread\_01" or "Thread\_US".
    \item \textit{goal}: The subtask execution target assigned to a subthread by the main thread.
    \item \textit{state}: Indicator of the current execution status of the thread, including "running," "successful," "failed," or "killed."
    \item \textit{allowed\_tools}: The list of tools that the subthread is permitted to use.
    \item \textit{prefix\_context}: The prefix context assigned to the subthread by the main thread, with its content and format determined entirely by the main thread.
    \item \textit{extra\_info}: Any additional remarks specified by the main thread, and this field is optional.
    \item \textit{start\_time}: The timestamp when the subthread is created and starts running, used to calculate and display its elapsed runtime.
    \item \textit{result}: The result returned by the subthread upon completion, corresponding to the task described in the goal field. This field is populated only after the subthread’s status is no longer "running."
\end{itemize}

\subsection{Inference}

All reasoning processes of the Self-Manager and the agent LLMs used in all baselines are implemented based on the SGLang framework \citep{zheng2024sglangefficientexecutionstructured}. All experiments are conducted on NVIDIA H100 GPUs. Both the tensor parallelism size and the expert parallelism size are set to 8. We use \texttt{torch.bfloat16} as the numerical precision for model weights, activations, and the KV cache during inference.

\section{Details of Experiment}
\label{appendix:experiment}

In this section, we present the implementation details of selected experiments and discuss other related experiments.


\subsection{Details of RQ3}

In RQ3, we investigate the issue of information retention. In context management modeling that supports long-horizon execution, any form of context transformation—such as summarization, folding, or the return of subthreads in a self-manager—may lead to the loss of critical information, which in turn degrades system performance. Therefore, it is necessary to design targeted experiments focusing on information retention to systematically examine this problem.

Specifically, based on the above problem formulation, we quantitatively evaluate information retention by invoking an LLM-as-Judge, as formalized in Equation \ref{eq:info_retention}. For a single execution trajectory, we focus on the positions where context transitions occur. At each such position, we prompt the LLM-as-Judge—using the template shown below—to extract a list of useful information units from the contexts before and after the transition, where usefulness is defined with respect to the task being solved in the current execution. We then invoke the LLM-as-Judge again to determine which information units from the pre-transition list are covered (matched) by the post-transition information units. A higher number of matched units indicates stronger information retention capability. Since ReAct does not involve any context transitions, it is excluded from this analysis. For ReSum, we compute information retention by comparing the contexts immediately before and after the agent loop invokes the summarization tool. For FoldAgent, we compare the branch execution context with the branch’s final answer, treating them as the pre- and post-transition contexts, respectively. For Self-Manager, we compare the context of a subthread during execution with the returned result after the subthread completes, which serve as the pre- and post-transition contexts.

\begin{tcolorbox}[breakable, colback=violet!10, colframe=violet, title=\textbf{Prompt Template for Information Extraction}]
You are a highly professional information extractor, skilled at identifying information points that are useful for a given task from long texts.\\\\
What Your Need to Do:\\
Given a task description and the contextual information produced by an Agent while completing that task, extract the information points from the context that are useful for the task.\\\\
Definition of an information point:\\
An information point should be a complete and unique statement of a fact (a sentence of about 10-20 words), and should include a clear subject, verb, and object, and, if necessary, constraint information such as time, location, topic, etc.\\\\
Definition of "useful":\\
If an information point is at least semantically relevant to the task description, it is considered useful.\\\\
Output Format:\\
- Output one plain-text nugget per line, with no other content.\\
- If no complete statement that is valuable to the task can be found in the passage, do not generate any low-quality nuggets, and simply return [None].\\
- Do not provide explanations, and ensure there is no redundant information.\\\\
Task Description: \textcolor{blue}{\monospace{\{task\_description\}}}\\
Context Information: \textcolor{blue}{\monospace{\{context\_info\}}}
\end{tcolorbox}

\begin{tcolorbox}[breakable, colback=violet!10, colframe=violet, title=\textbf{Prompt Template for Information Union}]
Task Description:\\
Given two Information Point Lists, A and B, you need to produce the union of the two Information Point Lists.\\\\
Definition of Information Point Lists:\\
- Each information point list contains multiple information points. Each information point is typically a complete and unique statement of a fact (a sentence of about 10-20 words).\\
- The number of information points in each list may vary.\\\\
Procedure:\\
1. Considering information points in List A, examine each information point one by one to determine whether it exists in List B. An information point is considered to exist as long as the fact it refers to is semantically supported by one or more information points in List B.\\
2. Output all information points from List A that are determined to exist in List B.\\\\
Output Format:\\
- One information point per line, in plain text. The output must correspond exactly to the information points in List A and remain completely unchanged.\\
- If none of the information points in List A are found to exist in List B, simply return [None].\\
- Do not provide explanations, and ensure there is no redundant information.\\\\
Information Point List A:\\
\textcolor{blue}{\monospace{\{info\_list\_a\}}}\\\\
Information Point List B:\\
\textcolor{blue}{\monospace{\{info\_list\_b\}}}
\end{tcolorbox}

\subsection{Details of RQ4}

For the computation of efficiency and cost, to ensure a fair comparison—by eliminating the influence of network APIs or local environment variability—we adopt a trajectory-based accounting approach, where all metrics are computed directly from the execution traces.

\paragraph{Time.} The agent runtime primarily consists of the time spent on base model invocations and tool calls. In FoldAgent, we assume that the folding operation—i.e., discarding a branch’s execution trace after the branch returns its result—does not incur significant overhead. Consequently, the runtime is dominated by base model calls and tool invocations. In Self-Manager, although multiple threads may execute in parallel, under the constraints of the framework, the start and end times of the main thread correspond to the earliest start and the latest completion among all threads. Therefore, we use the main-thread trajectory as the basis for runtime computation. For tool calls, we consider the time for thread creation, termination, and deletion to be negligible. However, when the main thread explicitly invokes a sleep operation to block itself, the specified sleep duration is counted toward the total runtime. Specifically, the unit-time costs of the operations considered in the runtime calculation are summarized in Table \ref{tb:time_rule}. The token generation time can be calculated based on the processing speed, as follows:
\begin{equation}
    \mathrm{Speed}=\frac{\mathrm{tokens}_{\mathrm{prompt} } + \mathrm{tokens}_{\mathrm{generation} } }{\mathrm{time} }  
\end{equation}

\begin{table}[h]
    \centering
    \caption{Unit runtime of agent execution, which is referenced from SGLang (\url{https://qwen.readthedocs.io/en/latest/getting_started/speed_benchmark.html\#qwen3-30b-a3b-sglang}). The context length limit is 128k. The quantization type is \texttt{torch.bfloat16}.}
    \begin{tabular}{cccc}
        \toprule
        \textbf{Item} & Speed (tokens/s) & $\text{tool}_{\text{{search}}}$ (s) & $\text{tool}_{\text{{visit}}}$ (s) \\
        \midrule
        \textbf{Time} & 1385.65 & 1.0 & 2.0 \\
        \bottomrule
    \end{tabular}
    \label{tb:time_rule}
\end{table}

\paragraph{Cost.} The agent execution cost is likewise dominated by base model invocations and tool calls. Although Self-Manager supports concurrency and asynchronous execution, it essentially renders base model calls asynchronous and parallel; as a result, the overall cost still primarily depends on the deployment cost of the base model. Specifically, the unit costs of the operations considered in the execution cost calculation are summarized in Table \ref{tb:cost_rule}.

\begin{table}[h]
    \centering
    \caption{Base pricing of agent runtime overhead. Token prices are based on Together.ai (\url{https://www.together.ai/pricing}), and search tool costs are based on Serper (\url{https://serper.dev/}).}
    \begin{tabular}{ccccc}
        \toprule
        \textbf{Item} & $\text{token}_{\text{{prompt}}}$ (\$/1M) & $\text{token}_{\text{{generation}}}$ (\$/1M) & $\text{tool}_{\text{{search}}}$ (\$/Call) & $\text{tool}_{\text{{visit}}}$ (\$/Call) \\
        \midrule
        \textbf{Price} & 0.80 & 0.80 & 0.001 & 0 \\
        \bottomrule
    \end{tabular}
    \label{tb:cost_rule}
\end{table}

\subsection{Discussion}

\begin{figure*}[h]
    \begin{minipage}[b]{0.47\textwidth}
    \centering
    \includegraphics[width=1.0\textwidth]{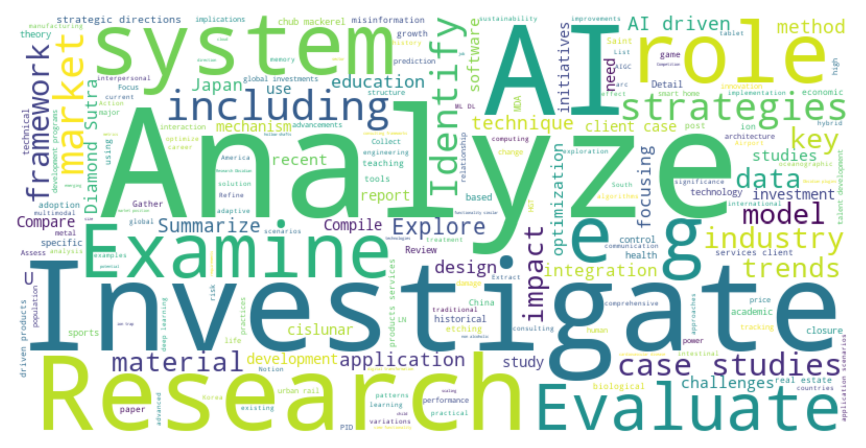}
    \caption{Word cloud of subthreads' subtasks assigned by the main thread in Self-Manager.}
    \label{fig:subtask_wordcloud}
    \end{minipage}
    \hfill
    \begin{minipage}[b]{0.50\textwidth}
    \centering
    \includegraphics[width=1.0\textwidth]{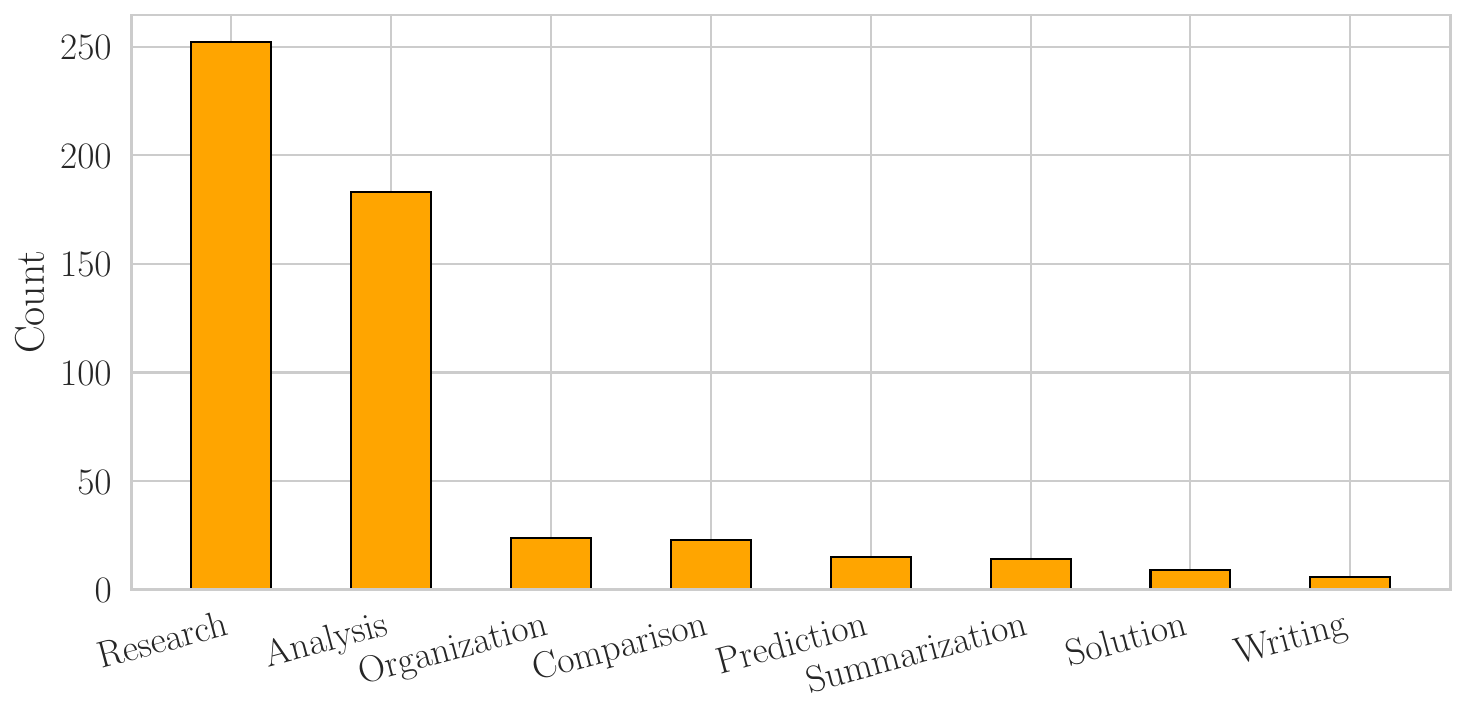}
    \caption{Frequency distribution of subtask types assigned to subthreads in Self-Manager.}
    \label{fig:subtask_bar}
    \end{minipage}
\end{figure*}

\paragraph{Subtask. } In Self-Manager, subtasks constitute a critical unit of processing. The main thread decomposes a complex task into multiple subtasks and can continuously discover new subtasks during execution. To enable independent execution, the main thread spawns child processes and dispatches subtasks to them. Accordingly, we investigate the behavior of the main thread when creating subtasks via spawning child threads. As illustrated in Figures \ref{fig:subtask_wordcloud} and \ref{fig:subtask_bar}, across a wide range of long-form deep research tasks, the types of subtasks assigned by the main thread span a diverse spectrum, including information gathering, problem analysis, material organization, and writing. Among these, research subtasks focused on information acquisition and analysis subtasks aimed at reasoning about and solving specific subproblems account for the majority. This observation is expected, as such subtasks are fundamental to long-form deep research: nearly any complex problem can be decomposed into research- and analysis-oriented subtasks. Meanwhile, we argue that models not explicitly trained to operate within a parallel agent loop are still limited in their ability to effectively leverage the highly flexible action of spawning child threads. A promising direction for future work is to train models to generate a broader and more diverse set of subtasks, thereby improving the utility and effectiveness of child threads.

\paragraph{Comparison with multi-agent systems. } Self-Manager falls within the category of single-agent systems. One of its core motivations is to address the generalization limitations inherent in workflow-based multi-agent systems. Generalization is precisely where single-agent approaches excel: under a universal design paradigm, a single agent can, in principle, execute any type of task in any scenario. Theoretically, an iteratively composable atomic loop is sufficient to construct the graph structure required to complete arbitrary tasks. In this work, Self-Manager extends the classical single-threaded agent loop by introducing concurrent and asynchronous subthreads. However, it does not cross into the multi-agent regime for the following reasons. (1) The system relies on a single agentic LLM. Although multiple threads execute different tasks concurrently, they all share the same policy. (2) The loop design remains universally applicable and can generalize to arbitrary tasks across diverse scenarios. Furthermore, inference frameworks such as SGLang can efficiently support concurrent multi-threaded requests to deployed models without introducing additional inference overhead. Consequently, by enabling self-management on subthreads, Self-Manager preserves the generalization advantages of single-agent inference while further mitigating issues present in existing agent loops.

\end{document}